\documentclass[11pt]{article}

\usepackage[preprint]{acl}

\usepackage{times}
\usepackage{latexsym}

\usepackage{times}
\usepackage{latexsym}
\usepackage{booktabs}
\usepackage{graphicx}
\usepackage{makecell}
\usepackage{tabularx}
\usepackage{times}
\usepackage{latexsym}
\usepackage{amsmath,amsfonts,bm}

\usepackage{enumitem}
\usepackage{multirow}
\usepackage{multicol}
\usepackage{xcolor}
\usepackage{wrapfig}

\usepackage[T1]{fontenc}

\usepackage[utf8]{inputenc}

\usepackage[utf8]{inputenc} 
\usepackage[T1]{fontenc}    
\usepackage{hyperref}       
\usepackage{url}            
\usepackage{booktabs}       
\usepackage{amsfonts}       
\usepackage{nicefrac}       
\usepackage{microtype}      
\usepackage{xcolor}         
\usepackage{colortbl}
\usepackage[most]{tcolorbox}
\usepackage{xspace}
\usepackage[table]{xcolor}
\usepackage{tabularx}

\newcommand{\DatasetName}{\textbf{OpenGenAlign}\xspace}
\usepackage{microtype}

\usepackage{inconsolata}

\usepackage{graphicx}

\title{OpenGenAlign: A Preference Dataset and Benchmark for Trustworthy Reward Modeling in Open-Ended, Long-Context Generation}

\author{
\textbf{Hanning Zhang\textsuperscript{1}\thanks{Work done during a research internship at NewsBreak.}},
\textbf{Juntong Song\textsuperscript{2}},
\textbf{Juno Zhu\textsuperscript{2}},
\textbf{Yuanhao Wu\textsuperscript{2}},
\textbf{Tong Zhang\textsuperscript{1}},
\textbf{Cheng Niu\textsuperscript{2}}
\\
 \textsuperscript{1}University of Illinois Urbana-Champaign,
 \textsuperscript{2}NewsBreak
\\
\texttt{\{hanning5, tozhang\}@illinois.edu}\\
  \texttt{\{juntong.song, juno, yuanhao.wu, cheng.niu\}@newsbreak.com}\\
}

\begin{document}
\maketitle
\begin{abstract}

Reward Modeling is critical in evaluating and improving the generation of Large Language Models (LLMs).
While numerous recent works have shown its feasibility in improving safety, helpfulness, reasoning, and instruction-following ability, its capability and generalization to open-ended long-context generation is still rarely explored.
In this paper, we introduce \DatasetName, a framework and a high-quality dataset designed to develop reward models to evaluate and improve \textit{hallucination-free, comprehensive, reliable, and efficient open-ended long-context generation}. We define four key metrics to assess generation quality and develop an automated pipeline to evaluate the outputs of multiple LLMs across long-context QA, Data-to-Text, and Summarization scenarios using o3, ending up with 33K high-quality preference data with a human agreement rate of 81\%.
Experimental results first demonstrate that existing reward models perform suboptimally on the held-out benchmark.
And Our trained reward model achieves superior performance in the benchmark and effectively improves the generation quality of the policy models using Reinforcement Learning (RL). Additionally, \DatasetName could be used for effective guided generation in existing datasets. Furthermore, we demonstrate that the \DatasetName could be integrated with reward data from other domains to achieve better performance\footnote{\url{https://github.com/hanningzhang/OpenGenAlign}}.

\end{abstract}

\section{Introduction}

Reward models have emerged as a critical component in aligning Large Language Models (LLMs) with human preferences, serving as the backbone of Reinforcement Learning from Human Feedback (RLHF) pipelines \citep{bai2022traininghelpfulharmlessassistant,lambert2024rewardbenchevaluatingrewardmodels,NIPS2017_d5e2c0ad,dong2024rlhfworkflowrewardmodeling,cui2025processreinforcementimplicitrewards}.
RLHF with explicit reward modeling has established itself as a standard paradigm for enhancing LLM capabilities and is now widely adopted across state-of-the-art models, including both proprietary systems such as GPT \citep{openai2024gpt4technicalreport} and Claude \citep{claude3_2024}, as well as open-source models such as Llama \citep{grattafiori2024llama3herdmodels}, Qwen \citep{yang2025qwen3}, and Gemma \citep{gemmateam2025gemma3technicalreport}.
In RLHF pipelines, reward models guide training by evaluating outputs and providing feedback signals for reinforcement learning optimization. This approach has proven effective across diverse objectives, from improving response helpfulness and safety \citep{bai2022traininghelpfulharmlessassistant,wang2024helpsteer2,liu2024skywork} to enhancing reasoning capabilities \citep{shao2024deepseekmathpushinglimitsmathematical,cui2025processreinforcementimplicitrewards,yuan2024advancingllmreasoninggeneralists}.

\begin{figure*}[htbp] 
    \centering
    \includegraphics[width=\linewidth]{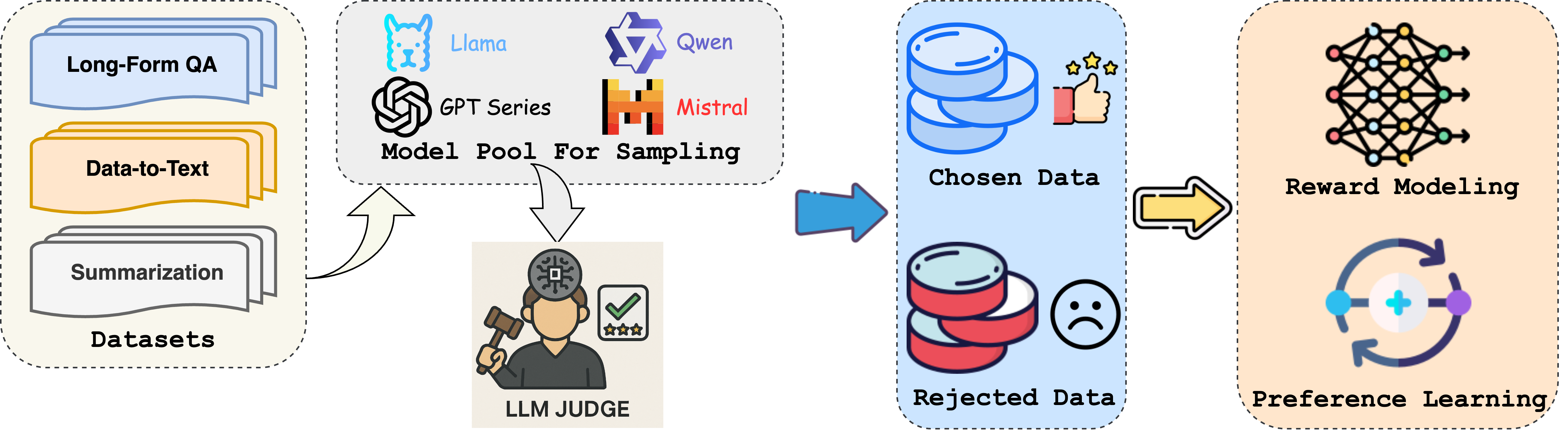} 
    \caption{Overview of our data labeling method and our experiments based on it in the open-ended long-context Scenario.
    We use o3 as the judge to evaluate the quality of the generation from multiple models.
    We then train the reward models and use them for Reinforcement Learning.} 
    \label{fig:overview} 
\end{figure*}


High-quality training data is fundamental to building effective reward models. Numerous datasets have been developed for this purpose, including HH-RLHF \citep{bai2022traininghelpfulharmlessassistant}, Ultra-Feedback \citep{cui2024ultrafeedbackboostinglanguagemodels}, Skywork-Reward \citep{liu2024skywork}, and HelpSteer-3 \citep{wang2025helpsteer3preferenceopenhumanannotatedpreference}. These datasets construct preference pairs by assessing response quality through either human annotation \citep{stienon2020learning,köpf2023openassistantconversationsdemocratizing} or frontier LLM evaluation \citep{cui2024ultrafeedbackboostinglanguagemodels,xu2024magpiealignmentdatasynthesis}, both of which have demonstrated effectiveness.
Correspondingly, various benchmarks have been developed to evaluate reward model performance, including RewardBench \citep{lambert2024rewardbenchevaluatingrewardmodels,malik2025rewardbench2advancingreward} and RM-Bench \citep{liu2024rmbench}. However, existing datasets and benchmarks leave a critical gap: there is neither comprehensive training data nor dedicated evaluation benchmarks for reward models in various open-ended, long-context generation settings.

While reward models have become standard, recent work has explored alternatives. Rule-based methods \citep{deepseekai2025deepseekr1incentivizingreasoningcapability,he2025skyworkopenreasoner1,wen2025lightr1curriculumsftdpo} and Reinforcement Learning with Verifiable Rewards (RLVR) \citep{lambert2025tulu3pushingfrontiers} bypass reward models by leveraging answer verification in domains like mathematical reasoning and code generation \citep{zeng2025simplerlzooinvestigatingtamingzero,zhang2025dpor1}. However, such approaches cannot extend to long-form generation, where the absence of ground truth and the subjective nature of quality assessment necessitate learned reward models to encode human preferences.


To address this gap, we introduce \DatasetName, a dataset for reward modeling in open-ended, long-context generation. \DatasetName comprises three components: a training set containing 33K preference pairs for reward modeling, a development set of 9K samples for policy optimization, and a held-out benchmark consisting of 1.5K samples for evaluating both existing reward models and our proposed model (Table~\ref{table:Datasets-stat}).
Specifically, we curate datasets spanning three domains: Question Answering, Data-to-Text, and Summarization.
Our data construction procedure operates as follows. We have a pool of 12 open-source and proprietary LLMs.
For each prompt, we randomly select two models from this pool to generate responses.
Subsequently, we employ OpenAI's o3 \citep{openai2025o3mini} model as the judge, augmented with majority voting across multiple comparisons, to assess response pairs according to four critical dimensions: \textbf{Hallucination}, \textbf{Comprehensiveness}, \textbf{Verbosity}, and \textbf{Attribution}.
This evaluation framework enables us to construct preference pairs—each comprising a chosen response and a rejected response. Figure~\ref{fig:overview} provides an overview of our complete pipeline.

Empirical results validate the effectiveness of our approach: reward models trained on \DatasetName achieve about 86\% accuracy on the held-out benchmark. Furthermore, we demonstrate that our reward model successfully guides policy optimization using the PPO algorithm \citep{schulman2017proximalpolicyoptimizationalgorithms}, showing substantial improvements in long-context generation quality. Our key contributions are summarized as follows:
\begin{itemize}
[leftmargin=*,label=$\bullet$,noitemsep,partopsep=0pt,topsep=0pt,parsep=0pt]
    \item We introduce a high-quality reward modeling dataset and benchmark for open-ended, long-context generation and release it to facilitate future research in this domain.
    \item We conduct comprehensive experiments demonstrating our reward model's effectiveness, including reward model training, policy training, out-of-distribution generalization, and integration with existing datasets.
\end{itemize}


\begin{table*}[ht]
\centering
\caption{Overview of several representative \textbf{Open-Sourced} preference datasets in chronological order used in reward model training. We compare \DatasetName with several popular datasets.}
\label{table:preference_datasets}
  \resizebox{0.95\textwidth}{!}{%
    \centering
\begin{tabular}{@{}lcc@{}}
\toprule
\textbf{Dataset} & \textbf{Size} & \textbf{Domain} \\
\midrule
OpenAI Summarization \citep{stienon2020learning}         & 93K      & Summarization\\
WebGPT-Comparisons \citep{nakano2021webgpt}  & 20K & General Web-based QA  \\
HH-RLHF  \citep{bai2022traininghelpfulharmlessassistant}                      & 161K      & Chat (Helpfulness, Harmlessness) \\
UltraFeedback \citep{cui2024ultrafeedbackboostinglanguagemodels}   & 64K    & Chat (Helpfulness, Honesty, Truthfulness, Instruction-Following)        \\
WildGuard \citep{wildguard2024} & 87K & Chat (Safety, Jailbreaks, Refusals) \\
UltraInteract  \citep{yuan2024advancingllmreasoninggeneralists}                 & 220K    & Reasoning (Math, Code) \\
HelpSteer2  \citep{wang2024helpsteer2}  & 10K   & Chat (Helpfulness, Correctness, Coherence, Verbosity, Complexity)  \\
Skywork-Reward-80K-v0.2  \citep{liu2024skywork}  & 80K   & Chat, Format Bias, Safety  \\
HelpSteer3  \citep{wang2025helpsteer3preferenceopenhumanannotatedpreference}  & 40K   & General Chat, Multi-lingual, Coding  \\ \midrule
\DatasetName & 33K & Open-ended, Long-form Generation \\
\bottomrule
\end{tabular}
}
\end{table*}

\section{Related Work}

\subsection{Reward Modeling and Reinforcement Learning}

Training reward models have become a widely used approach to align language models with human preference \citep{ouyang2022traininglanguagemodelsfollow}.
The alignment can both enhance their trustworthiness and helpfulness \citep{bai2022traininghelpfulharmlessassistant,wang2023helpsteermultiattributehelpfulnessdataset,cui2024ultrafeedbackboostinglanguagemodels}, and improve their problem-solving abilities \citep{dai2024processsupervisionguidedpolicyoptimization,yuan2024advancingllmreasoninggeneralists,zhang2024entropyregularizedprocessrewardmodel,cui2025processreinforcementimplicitrewards}.
Many high-quality datasets for reward modeling have been introduced, such as HH-RLHF \citep{bai2022traininghelpfulharmlessassistant}, Ultra-Feedback \citep{cui2024ultrafeedbackboostinglanguagemodels}, Code-UltraFeedback \citep{weyssow2024codeultrafeedbackllmasajudgedatasetaligning}, Ultra-Interact \citep{yuan2024advancingllmreasoninggeneralists}, HelpSteer \citep{wang2023helpsteermultiattributehelpfulnessdataset}, PKU-SafeRLHF \citep{ji2024pkusaferlhfmultilevelsafetyalignment}, Sky-Reward-Preference-80K \citep{liu2024skywork}, and HelpSteer-3 \citep{wang2025helpsteer3preferenceopenhumanannotatedpreference}.
The reward signal can be trained as a discriminative model to generate one or a few scalar values \citep{19ff28b9-64f9-3656-ba40-08326a05748e,liu2024skywork,wang2024interpretablepreferencesmultiobjectivereward}, or directly generated as next token prediction from language models \citep{zhang2024generativeverifiersrewardmodeling,zheng2023judgingllmasajudgemtbenchchatbot,xiong2024rlhflowmath}.
Inspired by Deepseek-R1 \citep{deepseekai2025deepseekr1incentivizingreasoningcapability}, there has been R1-like reasoning reward models which integrates long CoT and RL to further enhance the performance on various tasks \citep{chen2025rmr1rewardmodelingreasoning,liu2025inferencetimescalinggeneralistreward,xiong2025stepwiserstepwisegenerativejudges}.

Reinforcement Learning from human feedback (RLHF) is a widely used strategy to enhance policy models after the reward model is developed \citep{bai2022traininghelpfulharmlessassistant,kaufmann2024surveyreinforcementlearninghuman}.
RLHF plays a critical role in aligning LLMs with human values and achieving improved performance \citep{christiano2023deepreinforcementlearninghuman}.
Proximal Policy Optimization (PPO) is a commonly used algorithm for alignment tasks to enhance the policy models \citep{schulman2017proximalpolicyoptimizationalgorithms}. Alternative variants include Rejection Sampling Fine-tuning (RAFT) \citep{dong2023raftrewardrankedfinetuning}, Direct Preference Optimization (DPO) \citep{rafailov2024directpreferenceoptimizationlanguage}, Group Relative Policy Optimization (GRPO) \cite{shao2024deepseekmathpushinglimitsmathematical}, Group Sequence Policy Optimization (GSPO) \citep{zheng2025groupsequencepolicyoptimization}, and TreePO \citep{li2025treepobridginggappolicy}.
Rule-based reward without reward models has also been proven an effective way in reinforcement learning on verifiable domains \citep{zeng2025simplerlzooinvestigatingtamingzero,zhang2025dpor1,wen2025lightr1curriculumsftdpo}.


\subsection{Open-Ended Generation and Long Context Understanding for LLMs}

There has been much research towards the evaluation of the long context understanding, such as Needle-in-a-Haystack \citep{kamradt2023needle}, Ruler \citep{hsieh2024rulerwhatsrealcontext}, NeedleBench \citep{li2024needlebenchllmsretrievalreasoning}, where important pieces of information are placed inside long context and the LLMs are required to locate them. However, these benchmarks fail to evaluate the overall generation quality.
Other works, such as ZeroSCROLLS \citep{shaham2023zeroscrollszeroshotbenchmarklong}, LongBench \citep{bai2024longbenchbilingualmultitaskbenchmark}, and LongBench-v2 \citep{bai2025longbenchv2deeperunderstanding}, focus more on real-world scenarios like long Question-Answering and Summarization tasks. But they either limit the format in multiple-choice questions, or adopt metrics like F1 and ROUGE \citep{lin2004rouge}, which might not be completely accurate for long-form generation and might deviate from human judgment \citep{tan2024proxyqaalternativeframeworkevaluating}.
Early works \citep{liu2023gevalnlgevaluationusing,chiang2023largelanguagemodelsalternative} also attempt to incorporate LLMs as evaluators via prompting.
\citet{zhang2024longrewardimprovinglongcontextlarge} first evaluates long-context via LLMs, but only construct SFT and DPO data, without training any reward models.
And there are no available domain-specific datasets for effective training of such evaluators.

\begin{figure*} 
    \centering
    \includegraphics[width=0.9\linewidth]{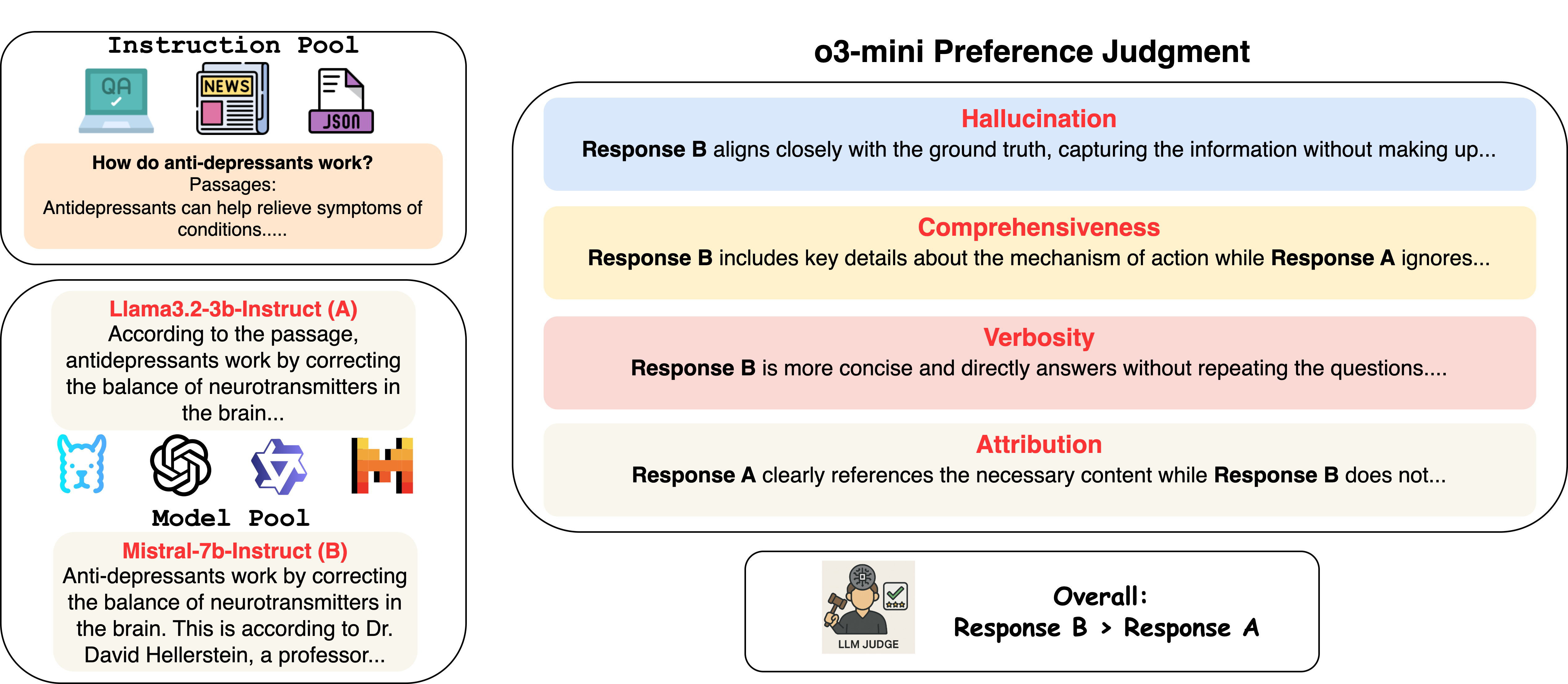} 
    \caption{Illustration of our data annotation method. Given a sample and two responses, we prompt o3 to provide judgments based on each metric and aggregate the results to construct pairs.} 
    \label{fig:annotation} 
\end{figure*}

\section{Dataset Construction}
\label{sec:data}

We construct our dataset based on existing open-ended long-context datasets to ensure its relevance and applicability.
To reflect the diverse use cases of different scenarios, we include three common types: Question Answering, Data-to-Text, and Summarization.
Specifically, we use WebGLM \citep{liu2023webglmefficientwebenhancedquestion}, Yelp \citep{Yelp}, and XSum \citep{narayan2018dontdetailsjustsummary} as experimental datasets, each dataset corresponding to one of the three scenarios introduced above.

\begin{table}
\vspace{-1.6em}
\centering
\caption{The number of preference pairs constructed from the three datasets used in our experiments.}
\label{table:we-use}
  \resizebox{0.44\textwidth}{!}{%
    \centering
\begin{tabular}{lccc}
\toprule
\textbf{Dataset} & \textbf{Training} & \textbf{Dev} & \textbf{Testing} \\
\midrule
WebGLM \citep{liu2023webglmefficientwebenhancedquestion} & 11,000 & 3,000 & 500 \\
Yelp \citep{Yelp}   & 11,000 & 3,000 & 500 \\
XSum \citep{narayan2018dontdetailsjustsummary} & 11,000 & 3,000 & 500 \\
\bottomrule
\end{tabular}
}
\end{table}

For the WebGLM dataset, LLMs are tasked with reasoning over web-retrieved reference data to answer real-world questions, generating concise responses in a few sentences.
For the Yelp dataset, our experiments focus on data from the restaurant category, represented in JSON format. Each sample includes information such as a restaurant's location and ambiance. Based on the structured JSON input, LLMs generate descriptive text about the restaurant.
The XSum dataset contains diverse articles from the British Broadcasting Corporation (BBC), with models tasked with summarizing these articles.
These three datasets cover a broad range of circumstances, ensuring that the reward model trained on them can significantly improve the development and evaluation of open-ended long-context understanding.
Table~\ref{table:we-use} presents the number of data samples used in our experiments. And examples of these data sets are presented in Table~\ref{table:Datasets-stat} Appendix.

When evaluating the quality of the responses, we consider the following metrics:
\paragraph{Hallucination:} The models should generate responses strictly based on the context provided, without introducing information not grounded in the context. If the context contradicts the model’s parametric knowledge, the model should adhere to the reference, ensuring that the response is accurate and contextually relevant \citep{niu2024ragtruthhallucinationcorpusdeveloping}.
\paragraph{Comprehensiveness:} The response should fully utilize the context provided by the retrieved content and address all aspects of the instruction \citep{zhu2024ragevalscenariospecificrag}. This requires the model to extract and integrate all relevant information from the retrieval context to ensure the response is thorough and complete.
\begin{figure*}[htbp] 
    \centering
    \includegraphics[width=\linewidth]{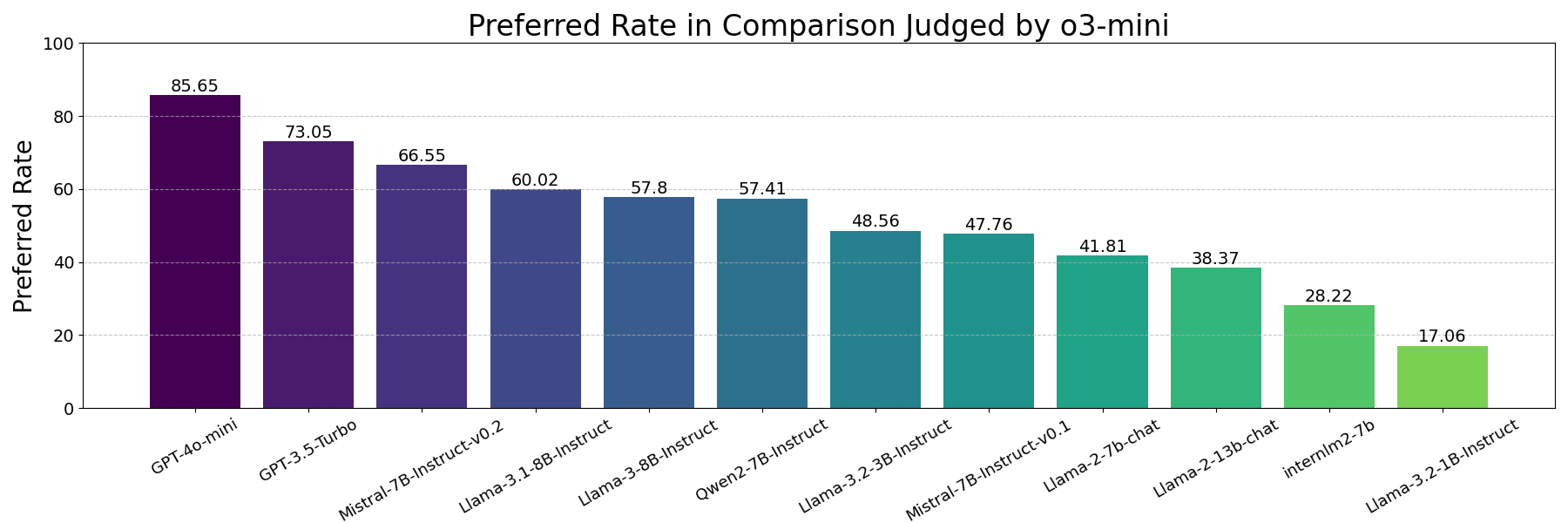} 
    \caption{Statistics of the preferred rate for each model during preference pairs construction phase.} 
    \label{fig:prefer-rate} 
\end{figure*}
\paragraph{Verbosity:} \citet{zhu2024ragevalscenariospecificrag} also adopts \textbf{Irrelevance} as a metric and we modify a bit. While the response should be detailed and comprehensive, it should also be concise, relevant, and straight to the point.
Striking the right balance between detail and brevity is essential to providing informative answers without overwhelming the user. This is especially important for the summarization task.

\paragraph{Attribution:} This metric is specifically applied to the WebGLM-QA dataset and this is important to ensure the generations are trustworthy and verifiable \citep{huang2024citationkeybuildingresponsible}.
The response should explicitly cite or point to the relevant part of the context when generating factual content.

\subsection{Dataset Sampling}
\label{sec:sampling}

We utilize a combination of open source instruction models, the GPT-3.5 \citep{brown2020languagemodelsfewshotlearners} and GPT-4 \citep{openai2024gpt4technicalreport} series to generate data, ensuring diversity and inclusion of both high-quality and relatively low-quality responses. The open-source models consist of various sizes of the instruction-tuned versions of Llama-3, Llama-3.1, Llama-3.2 \citep{grattafiori2024llama3herdmodels}, Llama-2 \citep{touvron2023llama2openfoundation}, Qwen-2 \citep{yang2024qwen2technicalreport}, InternLM-2 \citep{cai2024internlm2technicalreport}, and Mistral \citep{jiang2023mistral7b}.
In total, we include 12 candidate models for generation. For each question and its corresponding reference in the dataset, we randomly select two models' generations to form preference pair.

\subsection{Dataset Labeling}
\label{sec:labeling}

We have first compared the labeling quality of the reasoning model (o3) and non-reasoning model (GPT-4o), and identified the superior performance of the reasoning models. So we end up using o3 \citep{openai2025o3mini} to label the data.
An illustration of our labeling methods is shown in Figure~\ref{fig:annotation}.
Given a question $x$ and a pair of responses $(y_1,y_2)$ from different models, we prompt o3 to compare and select the preferred response.
Specifically, we ask o3 to compare the responses based on the four metrics outlined earlier, assessing them individually.
In the prompt, we explicitly ask o3 to put heavier weights on hallucination and comprehensiveness metrics, as they are crucial to the answer quality, while the other two mainly improves the readability.
After the o3 has made the individual judgments on the 4 metrics, it will generate an overall preference for the pair data based on the judgments above.
To increase reliability, we collect three independent judgments per question pair and use \textbf{majority voting} to determine the final preference.
And we end up getting the preference dataset of triplets $(x,y_w,y_l)$.
In our experiments, we observe no significant performance difference between evaluating attributes individually versus holistically in a single prompt; therefore, we adopt the latter approach to reduce computational costs. Our prompts for each dataset are shown in Figure~\ref{fig:webglm-prompt}.

\begin{table}
\vspace{-1.6em}
\centering
\caption{Human agreement rate of the \texttt{o3}-labeled dataset (measured as proportion).}
\label{table:eval-human-consistency}
  \resizebox{0.5\textwidth}{!}{%
    \centering
\begin{tabular}{lcccc}
\toprule
 & \textbf{WebGLM} & \textbf{Yelp} & \textbf{XSum} & \textbf{Avg} \\
\midrule
Agreement rate (\%) & 0.77 & 0.83 & 0.82 & 0.81 \\
\bottomrule
\end{tabular}
}
\end{table}


\begin{table*}[htbp]
\centering
\caption{The evaluation results of the existing reward models on the 3 tasks. They achieve SOTA performance on helpfulness and instruction-following, but do not excel in Long-Context Understanding.}
\resizebox{0.95\textwidth}{!}{%
\begin{tabular}{lcccc|cc}
\toprule
\textbf{Models}  & \textbf{WebGLM} & \textbf{Yelp} & \textbf{XSum} & \textbf{Average} & \textbf{RewardBench} \\ \midrule
\rowcolor{gray!10}
\textbf{Llama-3.1-8B-Instruct-RM-RB2} \citep{malik2025rewardbench2advancingreward} & \textbf{70.0} & \textbf{83.8} & \textbf{80.4} & \textbf{78.1} & 88.8 \\
\textbf{QRM-Gemma-2-27B} \citep{dorka2024quantile} & 67.4 & 83.2 & 77.4 & 76.0 & \textbf{94.4} \\
\rowcolor{gray!10}
\textbf{Skywork-Reward-Gemma-2-27B-v0.2} \citep{liu2024skywork} & 67.8 & 80.8 & 78.8 & 75.8 & 94.3 \\
\textbf{QRM-Llama3.1-8B-v2} \citep{dorka2024quantile} & 66.6 & 77.4 & 78.2 & 74.1 & 93.1 \\
\rowcolor{gray!10}
\textbf{URM-LLaMa-3.1-8B} \citep{lou2024uncertainty} & 64.6 & 83.6 & 73.0 & 73.7 &  92.9 \\
\textbf{Llama-3-OffsetBias-RM-8B} \citep{park2024offsetbias} & 65.6 & 78.2 & 77.4 & 73.7 & 89.4 \\
\rowcolor{gray!10}
\textbf{GRM-Llama3-8B-rewardmodel-ft} \citep{yang2024regularizing} & 66.6 & 74.8 & 79.4 & 73.6 & 90.9 \\
\textbf{Skywork-Reward-Llama-3.1-8B-v0.2} \citep{liu2024skywork} & 65.0 & 77.4 & 76.4 & 72.9 & 93.1 \\ \bottomrule
\end{tabular}
}
\label{table:eval-reward-benchmark}
\end{table*}

\subsection{Human Evaluation}
\label{sec:human-eval}

We also performed human evaluations to assess the alignment of AI annotations with human preferences.
To ensure the expertise and reliability, we hired \textbf{U.S. citizens with a major in Journalism} in college to compare and judge the quality of the dataset.
Specifically, we randomly select 100 samples with paired responses from each dataset and ask the annotators to evaluate using the same 4 metrics illustrated in Figure~\ref{fig:annotation}.
To ensure the consistency and quality of human annotation results, we ask 3 annotators to compare the pair and use majority voting results as the final preference.
The agreement ratio between the human annotators and o3 is shown in Table~\ref{table:eval-human-consistency}.
We observe an overall agreement rate of 81\%, with consistent agreement across the three tasks.
These results highlight proprietary LLMs' ability to effectively capture human preferences of subjective questions in assessing the response quality.

\section{Benchmark of Existing Reward Models}

In this section, we benchmark several existing reward models in our test set.
We select top models from RewardBench \citep{lambert2024rewardbenchevaluatingrewardmodels} known for their strong performance in assessing aspects such as helpfulness and instruction-following.
We examine their performance on diverse \DatasetName scenarios using our curated benchmark (test set) (Table~\ref{table:we-use}), and demonstrate their limitation in the evaluation on these domains.

The experiment results are shown in Table~\ref{table:eval-reward-benchmark}.
In addition to performance on the \DatasetName benchmark, we evaluate these models on RewardBench, which assesses their capabilities across chat, safety, and reasoning scenarios.
While all listed reward models achieve accuracy near or above 90\% on RewardBench domains, their performance drops below 80\% on \DatasetName, This underscores a significant gap between mainstream reward models and the unique requirements of tasks in \DatasetName, highlighting the distinct challenges posed by long-context generation tasks.
We further observe significant performance variation across tasks: reward models demonstrate stronger results on data-to-text and summarization tasks (e.g., Yelp and XSum datasets) compared to question-answering tasks (e.g., WebGLM dataset), indicating that current reward models lack uniform capability across different long-context scenarios.


Notably, we observe that strong performance on RewardBench does not guarantee comparable results on \DatasetName. For instance, Skywork-Reward-Llama-3.1-8B-v0.2 \citep{liu2024skywork} and URM-LLaMa-3.1-8B \citep{lou2024uncertainty} achieve relatively high overall scores on RewardBench but underperform on \DatasetName. Conversely, Llama-3.1-8B-Instruct-RM-RB2 \citep{malik2025rewardbench2advancingreward}, which scores lower on RewardBench, attains the highest accuracy on our benchmark. This disparity suggests that reward models optimized for general objectives (e.g., chat and safety) may not effectively generalize to long-context generation tasks, which require assessing different quality dimensions such as comprehensiveness and hallucination.

Overall, most of the existing reward models could not excel in expressing the preference in these scenarios.
Domain-specific training data are therefore essential to address this gap and improve long-context performance evaluation.

\section{Main Experiments}

We conduct both reward model training and reinforcement learning using our \DatasetName dataset. In total, 33K preference pairs are used for reward modeling (see Table~\ref{table:we-use}).
Additionally, we use a 9K-sample development set for sampling and learning during RLHF training. To evaluate the performance of the policy and reward models, a held-out benchmark set of 1.5K samples is used.

\subsection{Reward Modeling}
\label{Sec:reward-modeling}

We adopt the common approach to train the Bradley-Terry reward model \citep{19ff28b9-64f9-3656-ba40-08326a05748e,ouyang2022traininglanguagemodelsfollow} to learn the reward signal from the preference data.
Specifically, we use Llama-3.1-8B-Instruct \citep{grattafiori2024llama3herdmodels} as the base model for training.
We train the reward model with a learning rate of $2e^{-6}$, a global batch size of 64, a max length of 4096, and an epoch of 1 on 4 H100-80G GPUs. A formal formulation of the Bradley-Terry reward model is in Appendix~\ref{sec-math-formulation}.

\begin{table}
\centering
\caption{Evaluation results of the reward model on the three tasks. Accuracy is measured as the percentage of test samples where the reward model assigns a higher score to the chosen response than to the rejected one.}
\label{table:eval-reward}
  \resizebox{0.47\textwidth}{!}{%
    \centering
\begin{tabular}{lcccc}
\toprule
 & \textbf{WebGLM} & \textbf{Yelp} & \textbf{XSum} & \textbf{Avg} \\
\midrule
Accuracy (\%) & 80.2 & 92.0 & 85.6 & 85.9 \\
\bottomrule
\end{tabular}
}
\end{table}

\begin{table*}[h]
\centering
\caption{
\textbf{Win rates (\%)} of post-trained policy models on three tasks. Evaluation is done by both our reward model and \texttt{o3}, under two PPO settings: using our reward model vs. using the baseline reward model (Llama-3.1-8B-Instruct-RM-RB2).
}
\label{table:eval-rlhf-ours-vs-baseline}
\begin{tabular}{lcccccccc}
\toprule
\multirow{3}{*}{\textbf{Dataset}}
& \multicolumn{4}{c}{\textbf{Llama-3.1-8B-Instruct}}
& \multicolumn{4}{c}{\textbf{Llama-3.2-3B-Instruct}} \\
\cmidrule(lr){2-5} \cmidrule(lr){6-9}
& \multicolumn{2}{c}{Reward Model} & \multicolumn{2}{c}{\texttt{o3}}
& \multicolumn{2}{c}{Reward Model} & \multicolumn{2}{c}{\texttt{o3}} \\
\cmidrule(lr){2-3} \cmidrule(lr){4-5}
\cmidrule(lr){6-7} \cmidrule(lr){8-9}
 & Ours & Baseline & Ours & Baseline & Ours & Baseline & Ours & Baseline \\
\midrule
WebGLM & \textbf{89.6} & 75.8 & \textbf{85.2} & 66.6 & \textbf{83.6} & 70.8 & \textbf{72.4} & 62.6 \\
Yelp   & \textbf{91.6} & 75.6 & \textbf{89.4} & 74.1 & \textbf{95.6} & 86.4 & \textbf{91.0} & 85.0 \\
XSum   & \textbf{90.4} & 83.4 & \textbf{88.6} & 75.0 & \textbf{86.0} & 74.6 & \textbf{80.0} & 67.2 \\
\midrule
Average & \textbf{90.5} & 78.3 & \textbf{87.7} & 71.9 & \textbf{88.4} & 77.3 & \textbf{81.1} & 71.6 \\
\bottomrule
\end{tabular}
\end{table*}

We evaluate reward model performance using pairwise accuracy: for each test sample containing a chosen and rejected response, we measure whether the model assigns a higher score to the chosen response. As shown in Table~\ref{table:eval-reward}, our reward model achieves 85.9\% accuracy, demonstrating strong alignment with preference signals. Notably, our model outperforms all baselines from Table~\ref{table:eval-reward-benchmark}, achieving the highest accuracy and highlighting the benefits of training on domain-specific preference data for long-context generation.

Furthermore, we observe a consistent improvement across the 3 tasks, indicating that the reward model could jointly learn the preference signal from diverse tasks and domains.
Notably, the reward model achieves the highest accuracy on the Data-to-Text task, while its performance is relatively lower on the Question-Answering task.
This difference suggests that comparing structured data with text data is easier for the reward model, while evaluating the quality of a long-form QA poses a greater challenge. This observation aligns with our intuition and expectations.

\subsection{Reinforcement Learning}
\label{prefer-learning}

We adopt the Proximal Policy Optimization (PPO) algorithm \citep{schulman2017proximalpolicyoptimizationalgorithms} to perform the Reinforcement Learning.
We use Llama-3.2-3B-Instruct and Llama-3.1-8B-Instruct \citep{grattafiori2024llama3herdmodels} as the initial policy models
and use Verl \citep{Sheng_2025} as the framework for experiments.
For the baseline, we adopt Llama-3.1-8B-Instruct-RM-RB2 \citep{malik2025rewardbench2advancingreward}, which achieves the highest score in Table~\ref{table:eval-reward-benchmark}, to serve as the baseline reward model.
The training details and mathematical formulations are described in Appendix~\ref{sec:ppo}.

To quantify the effectiveness of alignment training, we evaluate our models using a \textbf{Win Rate} metric against the initial (pre-training) models. Specifically, we sample responses from a held-out test set using both the initial and post-trained policy models, generating paired responses for each prompt. These response pairs are evaluated through two methods: (1) scoring by the reward model, and (2) pairwise comparison by o3 based on the criteria defined in Section~\ref{sec:data}. For each evaluation method, we calculate the \textbf{win rate as the proportion of cases where the post-trained model's response is preferred over the initial model's response.} A win rate of 50\% represents no improvement.


The experiment results are shown in Table~\ref{table:eval-rlhf-ours-vs-baseline}. We observe a clear improvement in the policy models after PPO. Both the reward model and o3 agree that generations align more closely with the defined 4 metrics, as the average win rate is significantly above 50\%. These results highlight the effectiveness of our dataset and the reward model.
Additionally, the ratings across the 3 tasks from the reward model is very similar to the ratings from o3.
For example, our reward model and o3 both assign the highest preference on the Yelp dataset.
This shows that our reward model learns the rationale of rating from o3.
It is also observed that our reward model significantly outperforms the baseline when utilized for PPO, where the average win rate for Ours is about 10\% higher than the baseline reward model for PPO.
This comparison demonstrates the need for the \DatasetName, which could effectively improve the performance of both the reward model and policy model in these specific domains.
However, we also observe some imbalance in learning for the policy model across tasks.
Specifically, for the Llama-3.2-3B-Instruct model, the win rate for Yelp could reach 91\% while only above 70\% for WebGLM.
The comparison reveals that the difficulties are various across different domains in \DatasetName.
\begin{table*}

\centering
\caption{Performance of models in the guided generation on OOD datasets HotpotQA and MuSiQue.}
\label{table:guided-gen}
\resizebox{0.95\textwidth}{!}{%
\centering
\begin{tabular}{@{}lccc|ccc@{}}
\toprule
\textbf{Models} & \multicolumn{3}{c|}{\textbf{HotpotQA}} & \multicolumn{3}{c}{\textbf{MuSiQue}} \\
 & \textbf{PPO-Pass@1} & \textbf{Best-of-4@RM} & \textbf{Pass@1} & \textbf{PPO-Pass@1} &  \textbf{Best-of-4@RM} & \textbf{Pass@1}  \\
\midrule
\textbf{Llama-3.2-3B} & 37.2  & \textbf{39.3} & 28.5 & \textbf{15.3} & 15.0 & 8.7 \\
\textbf{Llama-3.1-8B}  & \textbf{48.1} & 46.4 & 38.3 & \textbf{26.4} & 24.9 & 14.6 \\
\bottomrule
\end{tabular}
}
\end{table*}

\begin{table*}[h]
\centering
\caption{Evaluation results of the Baseline and the MixReward Model, listing the accuracy of the reward models on each category and average from RewardBench \citep{lambert2024rewardbenchevaluatingrewardmodels} and \DatasetName test set.}
\begin{tabular}{l|ccccc|c}
\toprule
  & \textbf{Chat} & \textbf{Chat-Hard} & \textbf{Safety} & \textbf{Reasoning} & \textbf{Average} & \DatasetName  \\ \midrule
\textbf{Baseline} & 98.4 & 59.0 & 93.7 & 90.5 & 85.4 & 74.0 \\
\textbf{MixReward} & \textbf{98.7} & \textbf{59.2} & \textbf{94.0} & \textbf{92.4} & \textbf{86.1} & \textbf{86.0} \\ \bottomrule

\end{tabular}
\label{table:mix-reward}
\end{table*}



\subsection{Guided Generation}

We also evaluate our reward model on Out-Of-Distribution (OOD) multi-hop long-context Question-Answering datasets, including HotpotQA \citep{yang2018hotpotqadatasetdiverseexplainable} and MuSiQue \citep{trivedi2022musiquemultihopquestionssinglehop}.
We use the same policy model for generation.
Given a question, we first use the initial policy model to generate a single solution as the baseline, denoted as \textbf{Pass@1}.
Then we generate 4 solutions and apply the reward model to select the solution with the highest reward score, denoted as \textbf{Best-of-4}.
We also use the after-PPO model for evaluation, denoted as \textbf{PPO-Pass@1}
The experiment results are shown in Table~\ref{table:guided-gen}.
We observe a consistent and significant improvement for both the PPO models and \textbf{Best-of-4} compared to the baseline.
The results demonstrate the effectiveness and generalization of our reward model, as well as revealing the capability of the policy models when trained using \DatasetName.

\section{Training on Mixed Preference Datasets}

The ultimate goal of the proposed dataset is to further empower the reward modeling and provide a better judgment on the generation quality.
So a critical question is, how well could \DatasetName be integrated with existing preference datasets to build a more robust and versatile model.
We first construct the baseline dataset by randomly selecting 50K examples from Preference-700K\footnote{\url{https://huggingface.co/datasets/hendrydong/preference_700K}} \citep{dong2024rlhfworkflowrewardmodeling}, which is a mixture dataset consisting of Ultra-Feedback \citep{cui2024ultrafeedbackboostinglanguagemodels}, HH-RLHF \citep{bai2022traininghelpfulharmlessassistant}, PKU-SafeRLHF \citep{ji2024pkusaferlhfmultilevelsafetyalignment}, SHP \citep{ethayarajh2024ktomodelalignmentprospect}, HelpSteer \citep{wang2023helpsteermultiattributehelpfulnessdataset}, Ultra-Interact \citep{yuan2024advancingllmreasoninggeneralists}, Distilabel-Capybara \citep{distilabel-argilla-2024}, and Distilabel-Orca \citep{OpenOrca}, providing preference signals on chat, safety, and reasoning.
We train a baseline reward model using this dataset.
Additionally, we mix the baseline dataset with the \DatasetName to train new reward model named \textbf{MixReward}.
The training hyper-parameters and settings are the same as~\ref{Sec:reward-modeling}.
We evaluate the both the reward models on RewardBench \citep{lambert2024rewardbenchevaluatingrewardmodels} and our curated \DatasetName test set.

The experiment results are shown in Table~\ref{table:mix-reward}.
We first observe that, on the RewardBench, MixReward still possesses a competent performance compared to the Baseline, even with an about 1\% higher average accuracy, although we add 33K data sample irrelevant to chat, safety, or reasoning.
It demonstrates that \DatasetName could be effectively integrated with various existing preference datasets to train a more versatile reward model without hurting the performance on these domains.
Additionally, MixReward achieves an accuracy of 86.0\% on \DatasetName test set, which is significantly higher than the Baseline reward model.
Notably, the accuracy is very similar to the result in Table~\ref{table:eval-reward}, showcasing that adding datasets like Ultra-Feedback, SafeRLHF will also not hurt the performance of the reward model on \DatasetName domain.
These results together prove that \DatasetName could serve as a complementary resource to existing preference datasets, enhancing the overall diversity and strength of the resulting model, which also further demonstrates the quality and the effectiveness of our carefully curated dataset.

\section{Conclusion}

In this paper, we introduce \DatasetName, a high-quality preference dataset designed for evaluation and improvement of open-ended, long-context generation, which spans \textit{Question Answering, Data-to-Text, and Summarization} domains. Our dataset is generated through an automated AI annotation pipeline, leveraging both open-source and proprietary models to enhance generalization and versatility.
To ensure fair and reliable evaluations, we use o3 with majority voting to assess generation quality based on four key metrics carefully selected by human experts.
The experimental results show strong alignment with human evaluations, demonstrating the effectiveness of \DatasetName in reward modeling and reinforcement learning. These findings highlight the potential of our dataset to advance both the evaluation and generation of long-context scenarios. And the \DatasetName could be integrated into the training of a wide range of reward models. To foster further research, we will publicly released the dataset to the community.




\bibliography{custom}

\appendix
\clearpage
\newpage

\section{RLHF Experiment Details}
\label{sec-math-formulation}

\subsection{Bradley-Terry Reward Model}

To train the Bradley-Terry reward model, we utilize the preference dataset we collected.
We denote the dataset as $\mathcal{D} = (x, a^w, a^l)$, where $x$ is the prompt, $a^w$ is the preferred response, and $a^l$ is the dispreferred response.  After we get the dataset, we maximize the log-likelihood function of the BT model:

$$\mathcal{L}_{\mathcal{D}}(\theta) = \sum_{(x, a^w, a^l, y) \in \mathcal{D}} \log \left(\sigma \left(R_{\theta}(x, a^w) - R_{\theta}(x, a^l)\right)\right).$$

where $R_{\theta}$ is the output of the reward model, and $\sigma$ is the sigmoid function.

\subsection{Proximal Policy Optimization Experiments}
\label{sec:ppo}

Proximal Policy Optimization (PPO) is a reinforcement learning algorithm that has been popular in LLMs. It optimizes LLMs with the following objective function

$$
\mathcal{J}_{\text{PPO}}(\theta) = \mathbb{E}_{t,s_t,a_t\sim\pi_{\theta_{\text{old}}}}\left[\min\left(\frac{\pi_\theta(a_t|s_t)}{\pi_{\theta_{\text{old}}}(a_t|s_t)}\hat{A}_t, \text{clip}\left(\frac{\pi_\theta(a_t|s_t)}{\pi_{\theta_{\text{old}}}(a_t|s_t)}, 1-\epsilon_{\text{low}}, 1+\epsilon_{\text{high}}\right)\hat{A}_t\right)\right]
$$
where $\pi_\theta$ and $\pi_{\text{old}}$ represent the current and previous policy respectively. $\hat{A}_t$ is an estimator of the advantage function, and $\epsilon_{\text{low}}$ and $\epsilon_{\text{high}}$ are hyperparameters that control the maximum deviation from the previous policy $\pi_{\theta_{\text{old}}}$.

We perform PPO with a batch size of 512, max prompt length of 4096, max response length of 2048, and a learning rate of $1e^{-6}$.
We train the Llama3.1-8B-Instruct for 20 steps and the Llama3.2-3B-Instruct for 30 steps.

\begin{table*}[htbp]
\centering
\caption{Examples from the three datasets and the prompts used to construct preference data.
In WebGLM, LLMs generate answers based on reference passages; in Yelp, LLMs convert structured JSON to natural language descriptions; in XSum, LLMs summarize news articles. \textbf{Our Prompt} below demonstrates the prompt format we adopt to sample the response.}
\label{table:Datasets-stat}
\begin{tabularx}{\textwidth}{l|X}
\toprule
\textbf{Dataset} & \textbf{Data Example} \\
\midrule

\multirow{2}{*}{WebGLM \citep{liu2023webglmefficientwebenhancedquestion}}
& \makecell[l]{\textbf{\textit{Question:}} Why are different tiers (regular < mid < premium) of \\ gas' prices almost always 10 cents different? \\
\textbf{\textit{References:}} [The gap between premium and regular gas has..., \\ According to national averages, the price...] \\
\textbf{\textit{Answer:}} The 10 cent difference between the different tiers of gas \\ prices is likely due to a convention...}
\\ \addlinespace
& \makecell[l]{\textbf{\textit{Our Prompt:}} Answer the following question: \textit{\{question\}} \\
Your response should be based on the following passages: \textit{\{passages\}} \\
When you respond, you should refer to the source of information...}
\\
\midrule

\multirow{2}{*}{Yelp \citep{Yelp}}
& \makecell[l]{\textbf{\textit{Name:}} The Green Pheasant \quad \textbf{\textit{Address:}} 215 1st Ave S \\
\textbf{\textit{City:}} Nashville \quad \textbf{\textit{State:}} TN \\
\textbf{\textit{Attributes:}} \{ HappyHour: True, DogsAllowed: False, ... \}}
\\ \addlinespace
& \makecell[l]{\textbf{\textit{Our Prompt:}} Write an overview about the following business based \\ only on the provided structured data in the JSON format...}
\\
\midrule

\multirow{2}{*}{XSum \citep{narayan2018dontdetailsjustsummary}}
& \makecell[l]{\textbf{\textit{Document:}} The full cost of damage in Newton Stewart, one of the areas \\ worst affected, is still being assessed. Repair work is ongoing... \\
\textbf{\textit{Summary:}} Clean-up operations are continuing across the Scottish \\ Borders and Dumfries and Galloway after...}
\\ \addlinespace
& \makecell[l]{\textbf{\textit{Our Prompt:}} Summarize the following document: \textit{\{document\}}...}
\\
\bottomrule
\end{tabularx}
\end{table*}

\newpage
\begin{figure*}[t]
\caption{WebGLM System Prompt}
\label{fig:webglm-prompt}
\begin{tcolorbox}[title={WebGLM System Prompt}, enhanced, breakable]

 You are an expert in evaluating question-answering (QA) responses. Your task is to compare two responses, Response A and Response B, based on the following criteria, with a stronger emphasis on hallucination and comprehensiveness:
\newline \newline
\textbf{Primary Criteria (Most Important):
Hallucination (Faithfulness \& Accuracy) – Highest Priority}

Assess whether each response strictly adheres to the provided reference passages.
A good response must not introduce fabricated, misleading, or unverifiable information—it should only contain details supported by the reference.
Identify which response is more factually accurate and better grounded in the reference material.
\newline \newline
\textbf{Comprehensiveness (Coverage \& Relevance) – Second Priority}

Determine how well each response fully answers the given question, ensuring that all essential aspects are covered.
A strong response should capture key details from the reference while avoiding unnecessary or irrelevant information.
Identify which response provides a more complete and well-supported answer.
\newline \newline
\textbf{Secondary Criteria (Supporting Factors)
Conciseness (Efficiency \& Clarity)}

Evaluate how effectively each response conveys the necessary information without excessive verbosity.
The ideal response should be succinct yet informative, avoiding redundant details while preserving key insights.
If two responses are equally faithful and comprehensive, prefer the one that is more concise and well-structured.
\newline \newline
\textbf{Attribution (Attribution \& Use of Retrieved Content)}

Examine how well each response incorporates and attributes information from the retrieved passages.
A strong response should clearly reference relevant sources when necessary, ensuring that retrieved content supports the answer.
If two responses are otherwise equal, prefer the one that makes better citation of the retrieval sources.
\newline \newline
\textbf{Final Judgment:}
Focus primarily on faithfulness and comprehensiveness when deciding which response is superior. Provide a concise explanation of your reasoning, then explicitly state which response is better using the following format:

\textbf{Chosen: (A or B)}

\end{tcolorbox}
\end{figure*}

\newpage
\begin{figure*}[t]
\caption{XSum System Prompt}
\label{fig:xsum-prompt}
\begin{tcolorbox}[title={XSum System Prompt}, enhanced, breakable]

You are an expert in evaluating the quality of summarization. Your task is to compare two summaries, Response A and Response B, based on the following criteria:
\newline \newline
\textbf{Primary Criteria (Most Important): Hallucination (Faithfulness \& Accuracy)}

Assess how well each summary adheres strictly to the provided reference.
A good summary should only include verifiable information from the reference and avoid adding any fabricated, misleading, or exaggerated details.
Identify which summary is more factually accurate and better grounded in the reference.
\newline \newline
\textbf{Comprehensiveness (Coverage \& Relevance) – Second Priority:}

Determine how well each summary captures the key points of the reference without omitting essential details.
A strong summary should convey all critical aspects of the original content while avoiding irrelevant or unnecessary information.
Identify which summary provides a more complete and well-balanced representation of the reference.
\newline \newline
\textbf{Conciseness (Efficiency \& Clarity):}

Compare how effectively each summary delivers the key information in a compact and clear manner.
An ideal summary should be succinct yet informative, avoiding excessive verbosity while retaining all necessary details.
Determine which summary is more precise and effectively worded.
\newline \newline
\textbf{Final Judgment:}
Based on the above criteria, provide a brief explanation of your decision. Then, explicitly state which summary is the better one in the following format:

\textbf{Chosen: (A or B)}

\end{tcolorbox}
\end{figure*}

\newpage
\begin{figure*}[t]
\caption{Yelp System Prompt}
\label{fig:yelp-prompt}
\begin{tcolorbox}[title={Yelp System Prompt}, enhanced, breakable]

You are an expert in evaluating data-to-text generation. Your task is to compare two responses, Response A and Response B, which attempt to convert Yelp-style JSON data about a business into plain text. Evaluate them based on the following criteria:
\newline \newline
\textbf{Primary Criteria (Most Important)
Hallucination (Faithfulness \& Accuracy) – Highest Priority}

Assess whether each response strictly reflects the information provided in the JSON input.
A good response must not introduce fabricated, misleading, or unverifiable details—it should only include content supported by the JSON data.
Identify which response is more factually accurate and better grounded in the input.
\newline \newline
\textbf{Comprehensiveness (Coverage \& Relevance) – Second Priority}

Determine how well each response captures all the important details from the JSON (e.g., business name, category, rating, address, opening hours, reviews).
A strong response should present a comprehensive description while avoiding irrelevant or repeated details.
Identify which response provides a more complete and well-supported representation of the input.
\newline \newline
\textbf{Secondary Criteria
Conciseness (Efficiency \& Clarity)}

Assess how effectively each response presents the necessary information without being overly verbose.
The ideal response should strike a balance: concise enough to avoid redundancy, yet detailed enough to preserve key insights.
\newline \newline
\textbf{Final Judgment:}
Focus primarily on faithfulness and coverage when deciding which response is superior. Provide a concise explanation of your reasoning, then explicitly state which response is better using the following format:

\textbf{Chosen: (A or B)}

\end{tcolorbox}
\end{figure*}

\end{document}